\documentclass[10pt,twocolumn,letterpaper]{article}
\usepackage{multirow}
\usepackage{iccv}
\usepackage{times}
\usepackage{epsfig}
\usepackage{graphicx}
\usepackage{graphics}
\usepackage{adjustbox}
\usepackage{amsmath}
\usepackage{amssymb}
\usepackage{tabularx}
\usepackage{algorithm}
\usepackage{algorithmic}
\usepackage{paralist}
\usepackage[pagebackref=true,breaklinks=true,letterpaper=true,colorlinks,bookmarks=false]{hyperref}

\iccvfinalcopy 

\def\iccvPaperID{1613} 

\ificcvfinal\fi
\begin{document}

\title{
Low-bit Quantization of Neural Networks for Efficient Inference
}




\author{Yoni Choukroun
\qquad
Eli Kravchik
\qquad
Fan Yang
\qquad
Pavel Kisilev\\
Huawei Technologies Co.\\
\tt\small \{yoni.choukroun,eli.kravchik,yangfan74,pavel.kisilev\}@huawei.com
}

\maketitle

\begin{abstract}
Recent machine learning methods use increasingly large deep neural networks to achieve state of the art results in various tasks.
The gains in performance come at the cost of a substantial increase in computation and storage requirements. This makes real-time implementations on limited resources hardware a challenging task.
One popular approach to address this challenge is to perform low-bit precision computations via neural network quantization.
However, aggressive quantization generally entails a severe penalty in terms of accuracy, and often requires retraining of the network, or resorting to higher bit precision quantization.
In this paper, we formalize the linear quantization task as a Minimum Mean Squared Error (MMSE) problem for both weights and activations, allowing low-bit precision inference without the need for full network retraining.
The main contributions of our approach are the optimizations of the constrained MSE problem at each layer of the network, the hardware aware partitioning of the network parameters, and the use of multiple low precision quantized tensors for poorly approximated layers.
The proposed approach allows 4 bits integer (INT4) quantization for deployment of pretrained models on limited hardware resources. 
Multiple experiments on various network architectures show that the suggested method yields state of the art results with minimal loss of tasks accuracy.
\end{abstract}

\section{Introduction}

Neural networks (NNs) proved to be extremely effective in solving a broad variety of problems in computer vision, speech recognition and natural language processing \cite{krizhevsky2012imagenet,hinton2012deep,sutskever2014sequence}.
Deep learning methods are usually evaluated only according to their accuracy over a given task. This criterion leads to the development of architectures with constantly increasing computational complexity and memory requirements.
Thus, performing inference on low power System on a Chip (SoCs) used in smartphones or IoT devices is a significant challenge, due to the limited available memory and computational resources.  

Several approaches have been proposed in order to make deep NNs less resource demanding.
Network pruning of redundant and non-informative weights allows significant reduction of the network size \cite{han2015learning,han2015deep}.
Matrix factorization via low-rank approximation exploits the redundancy property of the NN parameters in order to increase speed up \cite{denton2014exploiting}. 
Distillation of NNs aims to transfer the knowledge contained in a pretrained large network to a compressed model via adapted training process \cite{hinton2015distilling}.
Also, new architectures (i.e.\cite{iandola2016squeezenet,howard2017mobilenets,huang2017densely}) with more efficient operations such as point/depth-wise or grouped convolutions allow the reduction of the model size, compared to existing over-parameterized architectures.

Another popular direction, which we focus on in this paper, is the quantization of NN. 
Quantization methods attempt to reduce the precision of the NN parameters and/or activations from single precision (32 bit floating point, or FP32) to lower bit representations.
Several benefits of low-bit precision can be exploited by deep learning accelerators. 
The storage requirement for a low-bit precision model can be diminished substantially, as well as the power consumption. 
Similarly, the memory bandwidth requirements can be significantly reduced.
Since the multiply accumulate (MAC) operations are performed on low-bit processing engines, the computational complexity can be reduced as well.
Perhaps the most important benefit of low bit representation is the saving of chip area. 
For instance, 8 bits integer (INT8) operations can save up to 30x energy and up to 116x area compared to FP32 operations \cite{dalli}, allowing significantly better computational throughput.
However, low-bit precision inference often causes loss of the task accuracy, which is usually compensated with the help of heavy full retraining, mixed precision or non-uniform quantization

In this paper, we address the quantization problem for weights and/or activations of a pretrained NN on highly constrained hardware, wherein  complete retraining or mixed precision calculations cannot be tolerated.
To the best of our knowledge, this is the first time \emph{INT4 only} deployment of a pretrained NN via efficient \emph{linear quantization} is performed with minimal loss of accuracy and data requirement.
We propose a simple yet efficient optimization framework to find the \emph{optimal} quantization parameters in the MMSE sense at each layer separately. 
The proposed MMSE reconstruction gives better accuracy than other existing MSE based optimization methods. 
NN parameters are quantized in a kernel-wise fashion that does not violate the linearity of the dot product operations, enabling efficient deployment on any common deep learning hardware.
We identify key layers that are most sensitive to quantization errors, and provide an adaptive quantization scheme that makes use of multiple low precision tensors.
Finally, we propose a refinement procedure for the scaling factors of the quantized tensors, which is performed on a small unlabelled calibration set.
In a similar manner, the NN activations quantization coefficients are obtained offline via MMSE criterion. More sensitive activations are better approximated according to their reconstruction residual.
The main contributions of this paper can be summarized as follows:
\begin{itemize}
\item We propose a low-bit precision \emph{linear} quantization framework for fast deployment of pretrained NNs on hardware which does not allow mixed precision operations.
\item We achieve minimal loss of task accuracy, while remaining compliant with modern deep learning hardware, by using fine grained partitioning of the NN weights. We propose optimal solution to the MMSE quantization problem and we deploy multiple tensors for key layers. Also, we refine the quantization factors of the network parameters for fast differentiable optimization. 
\item Extensive experiments on ImageNet with various popular architectures demonstrate that
our INT4 linear quantization method for both weights and activations, performs inference with only $3\%$ top-1 and $1.7\%$ top-5 mean accuracy degradation, as compared to the FP32 models, reaching, to the best of our knowledge, state-of-art results. The above degradation can be further reduced according to the complexity-accuracy trade-off inherent to the proposed multiple kernel method.
\end{itemize}
The remainder of the paper is organized as follows. Section 2 reviews  related works. In section 3, after analyzing the quantization challenges, we develop our MMSE based quantization for accurate approximation of original models. In this section we also provide experimental results to demonstrate the usefulness of the steps in the proposed quantization pipeline. At the end of the section, we describe the modification of the presented algorithm components for the problem of activation quantization.
Finally, quantization results from our experiments on several popular NNs are presented in Section 4.

\section{Related Work}
Neural network acceleration received increasing attention in the deep learning community, where the need for accurate yet fast and efficient frameworks is crucial for real-world applications.
A powerful approach is quantization of NNs to low-bit representation.
There are two main quantization scenarios. The first one is the full training of a given model to a desired lower bit precision. With this approach, the weights, the activations and even the gradients can be quantized to very low precision, enabling potentially fast training and inference \cite{zhou2016dorefa}.
The major problem with the training approach above arises from the discritness of the parameters, wherein the backpropagation approach is not well defined.
The "straight-through estimator" \cite{bengio2013estimating} has been used in \cite{courbariaux2015binaryconnect,rastegari2016xnor,courbariaux2016binarized} in order to estimate the gradient of a stochastic neuron.
\cite{courbariaux2015binaryconnect} proposed to use stochastic quantization of the weights via random rounding, in order to inject regularizing noise to the training process.
\cite{soudry2014expectation} suggests to approximate solution using variational Bayes method where the weights can be restricted to discrete values assuming Gaussian distribution.
Instead of seeking for appropriate derivatives, \cite{shayar2017learning} assumed smooth approximation of parameters with defined gradients.
Non-uniform quantization of NN parameters has been proposed in \cite{gong2014compressing} where the parameters are approximated using k-means algorithm.
\cite{li2017performance} proposed a high order quantization scheme of weights where the approximation residual is further processed allowing better refinement of full precision input.
Estimation of the quantization parameters by solving constrained optimization problem has been proposed for binary \cite{rastegari2016xnor} and ternary weights \cite{Li2016TernaryWN}.

We focus on the second quantization scenario that targets direct quantization of a pretrained FP32 network to a lower bit-depth precision without full training.
INT8 quantization of parameters has been proven to be relatively robust to quantization noise even with simple uniform quantization of weights \cite{jacob2017quantization}.
\cite{hwang2014fixed} proposed $L_2$ error minimization of weights via alternating optimization in order to obtain a generalizing ability during the training.
Nevertheless, INT8 quantization of the network activations is more challenging because of real time constraints.
Nvidia proposed in TensorRT \cite{migacz} a quantization framework that searches for saturation threshold of the activations, based on the Kullback-Leibler divergence measure between the quantized activations and their full precision counterpart.
Recently, \cite{banner2018aciq} proposed to approximate activations, as if they were sampled from a known distribution in order to obtain, under some assumptions, analytically optimal threshold in the $L_2$ sense.
However, quantization of full precision weights and activations to less than 8-bits usually causes significant loss of accuracy, a problem that has not been solved yet. 
In order to overcome such degradation in performance, quantization frameworks resort to retraining procedures, mixed precision solutions or non-uniform quantization. 
These solutions make fast and easy deployment of quantized NNs impossible, especially on highly constrained HW such as mobiles or IoT devices.

\section{Proposed MMSE Quantization Method}
Conversion of a full precision NN into its fixed point version introduces quantization noise (error) into the network.
Since significant noise may deteriorate the model performance, in the absence of training capabilities, efforts should be invested in minimizing the noise power, in order to approximate the original model as accurately as possible.
Minimization of the noise power of weights and of activations via MSE optimization is a natural criterion for quantization quality, even though no direct relationship can be easily established between the noise of the output and the model accuracy.
In the following, we investigate how quantization affects the NN output, and propose optimal solution to the quantization problem, via MSE optimization with fixed precision constraints.

\subsection{Quantization Process}
Consider an $L$-layer NN architecture defined by $\{W_l,X_l\}_{l=1}^{L}$ with $W_l$ and $X_l$ being the $l^{th}$ layer weight and input tensors respectively. 
Each weight is defined according to the set $W_l=\{W_{lk}\}_{k=1}^{k_l}$, where $W_{lk}\in \mathop{\mathbb{R}}^{c_l\times w_{kl}\times h_{kl}}$ denotes the $k^{th}$ kernel (neuron), with $c_l,w_{kl},h_{kl}$ being the number of channels, the kernel's width and height, respectively.
Similarly, $X_l\in \mathop{\mathbb{R}}^{c_l\times w_l \times h_l}$ represent the layer activations (feature maps), with $c_l,w_l,h_l$ being the number of channels, width and height of the layer input, respectively.

In our setting we constrain a tensor $T\in \mathbb{R}^{c\times w\times h}$ to be approximated \emph{linearly} as a quantized tensor $\tilde{T}\in \mathbb{Z}_p^{c\times w\times h}$ coupled with a scaling factor $\alpha\in\mathop{\mathbb{R}}$ such that $T\approx\hat{T}=\alpha\tilde{T}$. Here $p$ denotes the desired integer bit precision, and $\mathbb{Z}_p$ the corresponding quantization range.
Linear approximations are particularly appealing from hardware perspective, wherein the tensor multiplication unit can handle low precision tensors only. In this setting, the tensor multiplication is approximated as
\begin{equation}
\begin{aligned}
T_3 = T_1T_2\approx(\alpha_1\tilde{T}_1)(\alpha_2\tilde{T}_2)=(\alpha_1\alpha_2)(\tilde{T}_1\tilde{T}_2)=\alpha_3\tilde{T}_3.
\end{aligned}
\end{equation}

A popular uniform quantization scheme is given by
\begin{equation}
\begin{aligned}
\hat{T}& =\alpha\bigg[\frac{T-\delta}{\alpha}\bigg]_{\mathbb{Z}_p}+\delta\\
\delta &= \min_i (T_i)\\
\alpha &= \frac{\max_i(T_i)-\delta}{2^{p}-1}
\end{aligned}
\end{equation}
where $[T]_{\mathbb{Z}_p}= \min (\max (\lfloor T \rceil,\min (\mathbb{Z}_p)),\max (\mathbb{Z}_p))$ denotes the rounding operator $\lfloor \cdot \rceil$, followed by saturation to the $\mathbb{Z}_p$ domain.
The quantization offset $\delta$ can be of major importance for non symmetric distributions (e.g ReLU \cite{glorot2011deep} activations), where \emph{unsigned} representations $\mathbb{Z}_p=\mathbb{Z}_p^{+}\in\{0,\ldots,2^{p}-1\}$ allow an additional discrete level of representation. 
However, the use of offset obviously increases the computational complexity of tensor multiplications.
In the case quantization offset is not allowed, we assume $\delta=0$, $\alpha = \max_i(|T_i|)/(2^{p-1}-1)$ and the \emph{signed} range $\mathbb{Z}_p \in \{-2^{p-1},\ldots,2^{p-1}-1\}$. \\
Unless stated otherwise the results presented are obtained with signed quantization (no offset). 
Even though many quantization methods do not quantize the first and last layer of the model \cite{han2015learning,zhou2016dorefa,rastegari2016xnor}, unless stated otherwise here we quantize \emph{all} the network parameters and activations (including the network input).

\subsection{Mean Squared Error Analysis of Quantization}
The relation between the full precision tensor weights $W$ and activations $X$ and their respective approximations $\hat{W}$ and $\hat{X}$ can be obtained as follows
\begin{equation}
\begin{aligned}
\hat{Y}&=\hat{W}\hat{X}=(W+n_W)(X+n_X)\\
&=WX+Wn_x+n_WX+n_Wn_X\\
&\approx WX+Wn_X+n_WX=Y+Wn_X+n_WX,
\end{aligned}
\end{equation}
where $n_W,n_X$ denote the quantization noise of the weights and activations respectively and where the approximation is obtained by neglecting second order noise term.
Let us consider the case where the NN is composed of linear layers only. In such setting the NN output $Y$ is defined as
\begin{equation}
	Y=W_L(W_{L-1}...(W_1X_1))=W_LX_L.
\end{equation}
For ease of notation we will omit the mean factor of the MSE. 
Defining $e_L^2$, the MSE between the original model output and the quantized model output, we obtain in expectation that
\begin{equation}\label{eq_recurs}
\begin{aligned}
	\mathop{\mathbb{E}}(e_L^2)&=\mathop{\mathbb{E}}\|Y-\hat{Y}\|^2_F=\mathop{\mathbb{E}}\|W_LX_L-\hat{W}_L\hat{X}_L\|_F^2\\
	&=\mathop{\mathbb{E}}\|W_Ln_{X_L}+n_{W_L}X_L\|_F^2\\
	&=\mathop{\mathbb{E}}\|W_Ln_{X_L}\|^2_F+2\mathop{\mathbb{E}}trace(n_{X_L}^{T}W_L^{T}X_Ln_{W_L})\\
	&+\mathop{\mathbb{E}}\|n_{W_L}X_{L}\|_F^2\\
	&\approx\mathop{\mathbb{E}}\|W_Ln_{X_L}\|^2_F+\mathop{\mathbb{E}}\|n_{W_L}X_L\|_F^2\\
	&\leq \mathop{\mathbb{E}}\|n_{W_L}\|_F^2\mathop{\mathbb{E}}\|X_L\|_F^2  +\mathop{\mathbb{E}}\|W_L\|_F^2\mathop{\mathbb{E}}\|n_{X_L}\|^2_F\\
	&=\mathop{\mathbb{E}}\|n_{W_L}\|_F^2\mathop{\mathbb{E}}\|X_L\|_F^2  +\mathop{\mathbb{E}}\|W_L\|_F^2\mathop{\mathbb{E}}(e_{L-1}^2),
\end{aligned}
\end{equation}
where the approximation is obtained by assuming zero mean noise \cite{polino2018model} and where $\|\cdot\|_F$ denotes the Frobenius norm.
The inequality is obtained using Cauchy Schwarz inequality and by assuming the weights/activations and the noises are statistically independent. 
We obtain here a recursive expression of the network output MSE. It is obvious that first layers may have significant impact on the output quality because of the recursion factor.
Also, special attention should be given to the minimization of the weights noise which is coupled with possibly unbounded activations (ReLU).
In real configurations, the non linear functions that play major role in the discriminative power of deep networks are much harder to model and analyze.
Nevertheless, the linear analysis presented above, is supported empirically in our experiments, wherein the first layers always impact model accuracy the most.
Also, in our setting, the approximation quality of the weights has much more influence on the performance than the activations, in contrast to the analysis of \cite{zhou2016dorefa}.
\subsection{Kernel-Wise Quantization}
In the low precision set up, such as 4 bit, \emph{global} quantization where all the quantized kernels are coupled with a \emph{single} scaling factor, such that $\hat{W}_l=\{\alpha_l\tilde{W}_{lk}\}_k$, can lead to poor performance, due to the high variance of all the accumulated kernel elements. This problem does not arise in the INT8 deployment setting that is robust enough even under global quantization \cite{migacz,jacob2017quantization}.
Also, applying different scaling factors to different buckets (partitions) of contiguous values allows better approximation of the original tensor \cite{alistarh2017qsgd}. 
However, the quantization framework should preserve the linearity of the dot product, so that for every two vectors $x,y \in \mathbb{R}^n$ we get
\begin{equation}
\begin{aligned}
\langle x,y\rangle \approx \langle \alpha \tilde{x},\beta \tilde{y} \rangle= \sum_i^n\alpha \beta \tilde{x}_i\tilde{y}_i=\alpha \beta \langle\tilde{x},\tilde{y} \rangle.
\end{aligned}
\end{equation}
Maintaining the linearity is not so popular because of its restrictiveness and sub-optimality. For example, with the efficient group-wise \cite{alistarh2017qsgd,mellempudi2017ternary}, channel-wise or pixel-wise quantization settings \cite{faraone2018syq}, the dot product must be split into several dot products to be accumulated together.
Such procedures cannot be efficiently implemented on deep learning accelerators with dedicated matrix multiplication units such as systolic arrays \cite{jouppi2017datacenter}.
The only viable bucketing approach is the \emph{kernel wise} approach where each kernel is coupled to its own scaling factor such that $\hat{W}_l=\{\alpha_{lk}\tilde{W}_{lk}\}_k$. 
Such setting maintains the dot product linearity at the kernel level and allows efficient tensors multiplication in low precision dedicated hardware. 
Each discrete feature map is then multiplied with its corresponding floating point scaling factor.
Thus, without violating the dot product linearity and with negligible storage addition (as the number of kernels), a significant improvement in performance can be obtained as summarized in Table \ref{table:kernelwise}.
In this experiment, all convolutional layers are quantized \emph{kernel-wise}, while all the fully connected layers are \emph{globally} quantized.
Also, we only allow \emph{global} quantization of activations since every possible partitioning yields separation of the dot product induced by the convolutions. 
\begin{table}
\begin{center}
\begin{tabular}{|l|c|c|c|}
\hline
Architecture & Original &Global & Kernel-wise\\
\hline\hline
Alexnet \cite{krizhevsky2012imagenet} & 56.624\%      & 0.694\% & \textbf{46.796}\%\\
VGG16bn \cite{simonyan2014very}& 73.476\%      & 3.034\% & \textbf{65.23}\%\\
Inception v3 \cite{szegedy2016rethinking} & 76.226\% & 0.106\% & \textbf{12.564}\%\\
Resnet18 \cite{he2016deep}& 69.644\%     & 1.83\%  & \textbf{44.082}\%\\
Resnet50 \cite{he2016deep}& 76.012\%     & 0.172\% & \textbf{62.242}\%\\
Resnet101 \cite{he2016deep}& 77.314\%     & 0.148\% & \textbf{64.434}\%\\
SqueezeNet \cite{iandola2016squeezenet}& 58.0\%     & 1.528\% & \textbf{29.908}\%\\
DenseNet \cite{huang2017densely}& 74.472\%     & 0.58\%  & \textbf{57.072}\%\\
\hline
\end{tabular}
\end{center}
\caption{Top1 accuracy using INT4 quantization of weights (FP32 activations) with global scaling factor for all kernels and the proposed kernel-wise approach on the ImageNet validation set. Global quantization simply collapses.}
\label{table:kernelwise}
\end{table}

\subsection{Minimum MSE Quantization}\label{section:optimal_mse}

\begin{figure}[t]
\begin{center}
   \includegraphics[width=1\linewidth]{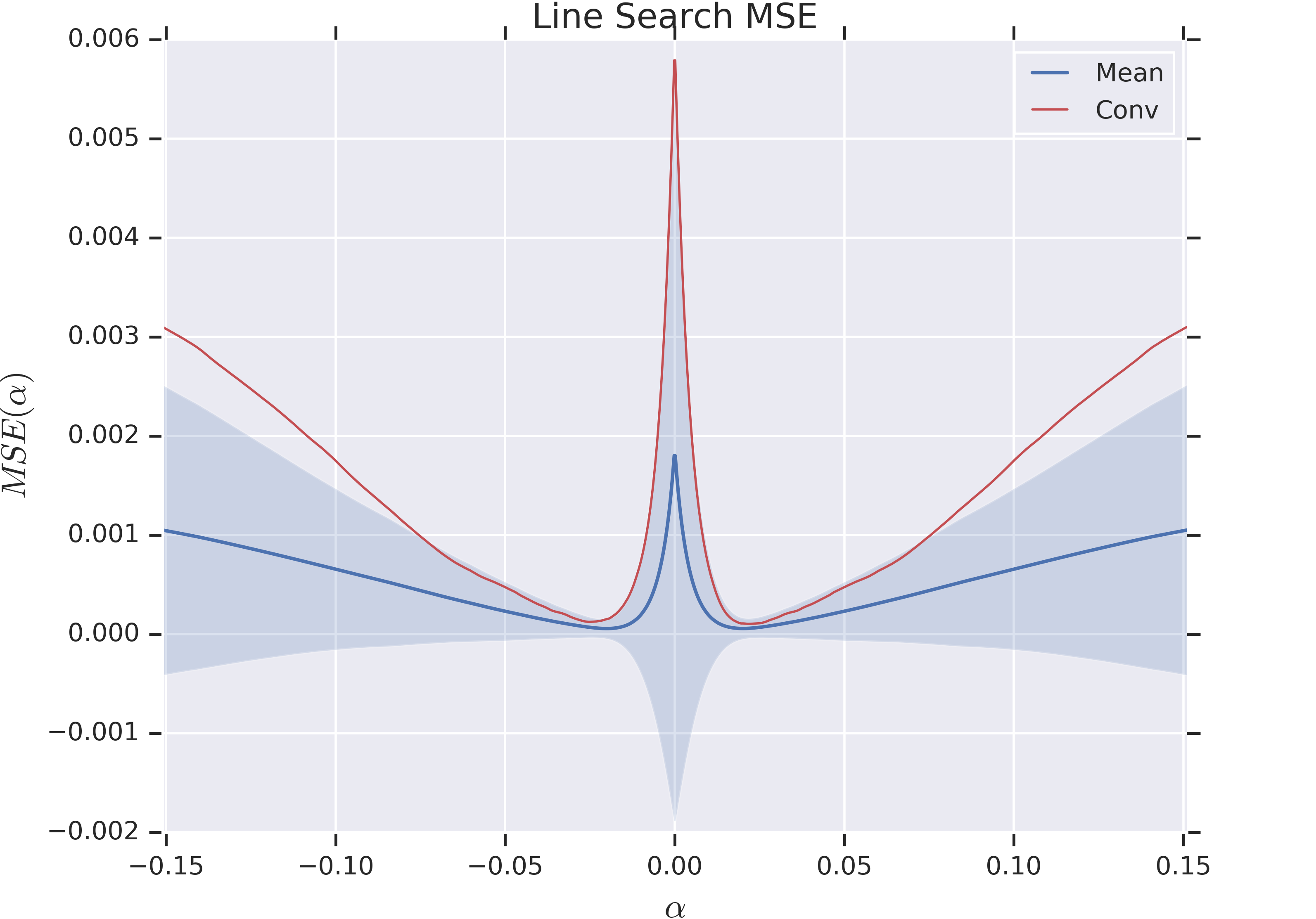}
\end{center}
   \caption{MSE as a function of the scaling factor $\alpha$ averaged for all the Alexnet convolutional kernels (Mean). Typical kernel is presented in red (Conv) to show the non-convexity of the function.
   }
   \label{fig:1D_LS}
\end{figure}

We aim at finding optimal approximation $\hat{T}$ of a given tensor $T\in \mathbb{R}^{c\times w\times h}$ by solving the constrained MSE problem as follows
\begin{equation}
\label{eq:opt_mse}
\begin{aligned}
& \underset{\alpha,\tilde{T}}{\text{min}}
& & \|T-\alpha\tilde{T}\|^2_F. \\
& \text{s.t}
& & \alpha \in \mathop{\mathbb{R}}, \tilde{T}\in \mathbb{Z}_p^{c\times w\times h}
\end{aligned}
\end{equation}
For any arbitrary precision $p$ this optimization problem does not have an analytical solution.
Assuming optimal $\alpha\neq0$ is given, and denoting $T_i$ the $i^{th}$ element of the tensor $T$, we remain with a one constraint optimization problem where 
we can rewrite eq. (\ref{eq:opt_mse}) objective as
\begin{equation} \label{eq:linesearch}
\begin{aligned}
\|T-\alpha\tilde{T}\|^2_F = \sum_{i}(T_i-\alpha \tilde{T}_i)^2=\alpha^2\sum_{i}(\frac{T_i}{\alpha} -\tilde{T}_i)^2.
\end{aligned}
\end{equation}
Thus, the \emph{optimal} quantization is uniform and the quantized values are given according to
$\tilde{T_i}=\big[T_i/\alpha\big]_{\mathbb{Z}_p} \forall i$,
where the scaling factor $\alpha$ also defines \emph{saturation} of the the tensor values.
Typical MSE as a function of $\alpha$ is presented in Figure \ref{fig:1D_LS}.
Several methods for finding optimal $\alpha$ can be considered.
Alternating optimization \cite{hwang2014fixed}, even on one half of the domain, is not robust due to
the non-convexity of the quantization operation (Figure \ref{fig:1D_LS}).
Alternatively, Golden Section Search gives fast and better results because of the low resolution optimization at the early stage of the algorithm, that avoids attraction to local minima.
In this work, we propose to use one dimensional exact line-search that can be implemented very efficiently over a given grid (few hundred points) using parallel computation. This approach allows \emph{optimal} solution (up to the grid density) of the MMSE quantization problem.
A comparison of different quantization methods is presented in Table \ref{table:omse}.

\begin{table}
\begin{center}
\begin{tabular}{|p{1.75cm}|p{1.1cm}|p{1.1cm}|p{1.1cm}|p{1.1cm}|}
\hline
Architecture  	&Uniform 		& Altern. 		& Golden				& \ OMSE \\
\hline\hline
Alexnet 			& 46.796\% 		& 40.962\%		& 46.070\%			& \textbf{46.892}\% \\
VGG16bn 			& 65.23\%  		& 55.936\%		& 62.250\%			& \textbf{65.414}\% \\
Inception v3 	& 12.564\%		 & 4.368\% 		& 7.408\%			& \textbf{22.028}\%\\
Resnet18 		& 44.082\% 		& 52.646\%		& 52.398\%			& \textbf{56.688}\%\\
Resnet50 		& 62.242\% 		& 60.186\%		& 63.178\%			& \textbf{67.356}\%\\
Resnet101 		& 64.434\% 		& 56.282\%		& 63.052\%			& \textbf{65.066}\%\\
SqueezeNet 		& 29.908\% 		& 24.040\%		& \textbf{34.036}\%	& 32.262\%\\
DenseNet 		& 57.072\% 		& 45.668\% 		& 50.058\%			& \textbf{59.012}\%\\
\hline
\end{tabular}
\end{center}
\caption{Top1 accuracy for INT4 quantized weights and FP32 activations using uniform quantization, Alternating optimization (Altern.)\cite{hwang2014fixed}, Golden Section search (Golden) and the proposed optimal MSE (OMSE) quantization.}
\label{table:omse}
\end{table}
\subsection{Multiple Tensor Quantizations}
\label{section:multiple}
Some layers can be harder to approximate than others, and have larger impact on the model accuracy loss.
In order to better reconstruct those layers, some works use mixed precision \cite{lin2016fixed}, where some layers use more bits for representation.
Due to power and chip area considerations, low power devices generally do not allow mixed precision inference, because of the constrained low precision engines. We therefore propose to better approximate \emph{key} layers by using multiple low precision quantized tensors. 
We use the term key layers to describe quantized layers with \emph{high MSE}, or MSE bigger than some threshold $\tau$.
In contrary to other training methods where kernels are blindly added over all the network layers \cite{lin2017towards}, our approach allows \emph{minimal} computational overhead.
We formalize this setting as follows.
Given $n$ integers $\{p_i\}_{i=1}^{n}$ denoting the desired precision of the quantized tensors and a given tensor $T$, we want to find the $n$ scalars $\{\alpha_i\}_{i=1}^{n}$ and the quantized tensors $\{\tilde{T}^i\}_{i=1}^{n}$ that best approximate the original tensor in the $L_2$ sense, such that
\begin{equation}
\begin{aligned}
& \underset{\alpha_i,\tilde{T^i}}{\text{min},}
& & \|T-\sum_{i=1}^{n}\alpha_i\tilde{T^i}\|^2_F \\
& \text{s.t}
& & \alpha_i \in \mathop{\mathbb{R}}, \tilde{T}^i\in \mathbb{Z}_{p_i}^{c\times w\times h}
\end{aligned}
\end{equation}
From the computational perspective, the proposed framework approximates a given kernel with quantized filter bank for better approximation. Thus, for a given tensor $X$ approximated as $X\approx\hat{X}=\beta\tilde{X}$ we have
\begin{equation}
\begin{aligned}
TX\approx\sum_{i=1}^{n}\alpha_i\tilde{T^i}X\approx\sum_{i=1}^{n}\alpha_i\beta(\tilde{T^i}\tilde{X}).
\end{aligned}
\end{equation}
Here we opt for a nested optimization approach, where the solution is obtained iteratively until convergence.
Then, the optimization can be written in the nested form as
\begin{equation}
\begin{aligned}
& \underset{\alpha_1,\tilde{T}^1}{\text{min} }\{ \underset{\alpha_2,\tilde{T}^2}{\text{min} } ...\{\underset{\alpha_n,\tilde{T}^n}{\text{min}\{ }
& & \|T-\sum_{i=1}^{n}\alpha_i\tilde{T}^i\|^2_F\} \}\}
\end{aligned}
\end{equation}
Alternating optimization is used until convergence, while at each iteration the quantized tensor and its scaling factor are obtained using the MSE quantization mapping suggested in Section \ref{section:optimal_mse}.
The proposed quantization approach for layers with high MSE is summarized in Algorithm \ref{alg:dual}. In the algorithm below, input $T$ refers to one convolutional kernel of a given key layer.
 \begin{algorithm}[H]
 \caption{Alternating Optimization for Multiple Quantization of high MSE layer}
 \label{alg:dual}
 \begin{algorithmic}[1]
 \renewcommand{\algorithmicrequire}{\textbf{Input:}}
 \renewcommand{\algorithmicensure}{\textbf{Output:}}
 \REQUIRE Tensor $T$, desired precision $\{p_i\}_{i=1}^{n}$ and quantization mapping $\phi$.
 \ENSURE  $\{\alpha_i\}_{i=1}^{n}$ and $\{\tilde{T}_i\}_{i=1}^{n}$
  \WHILE{convergence rate $>\epsilon$}
     \FOR {$j \in [1,...,n]$}
     \STATE $(\alpha_j,\tilde{T}^j)=\phi(T-\sum_{i=1,i\neq j}^{n}\alpha_i\tilde{T}^i,p_{j})$
     \ENDFOR
  \ENDWHILE
 \end{algorithmic}
 \end{algorithm}

\begin{figure}[t]
\begin{center}
   \includegraphics[width=\linewidth]{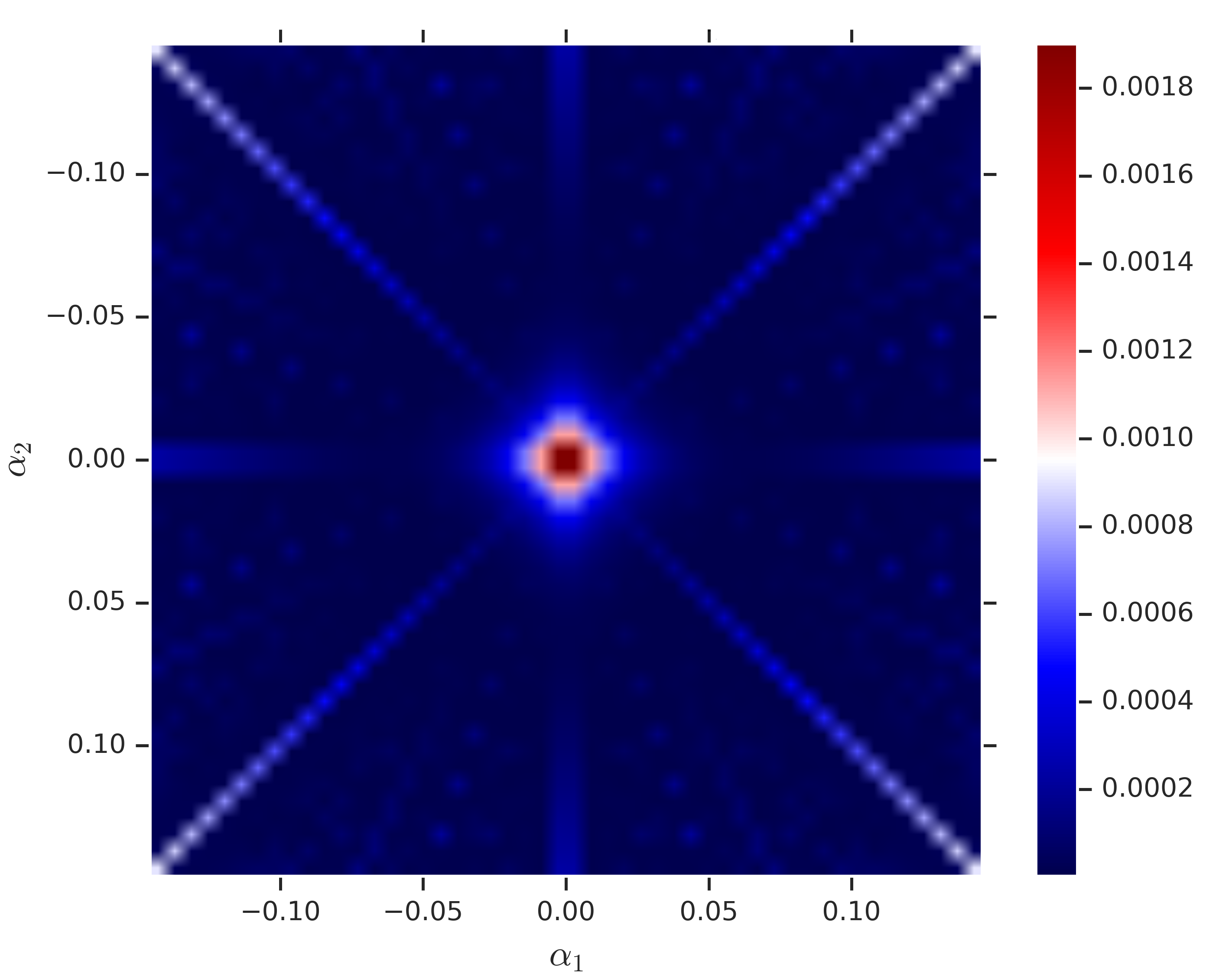}
\end{center}
   \caption{Landscape of the Dual MSE objective as a function of the scaling factors $\alpha_1$ and $\alpha_2$ (MSE averaged over all the Alexnet convolutional kernels).}
   \label{fig:2D_LS}
\end{figure}

For the special \emph{dual} case where $n=2$, optimum can be obtained efficiently by resorting to dual line search. 
Since each element in $\tilde{T}^1$ can only have discrete values in $\mathbb{Z}_{p_1}$, one can evaluate $\tilde{T}^2$ for all the few $2^{p_1}$ allowed quantized values and later select the integer value giving minimum of the objective. 
Assuming $\alpha_1,\alpha_2$ are given via grid search,
we want to find $\forall j$
\begin{equation}
\begin{aligned}
& \underset{\tilde{T}^1_j,\tilde{T}^2_j}{\text{min} }
& &(T_j-\alpha_1\tilde{T}_j^1-\alpha_2\tilde{T}_j^2)^2\\
\iff & \underset{\tilde{t} \in \mathbb{Z}_{p_1}}{\text{min} } \ \ \ \underset{\tilde{T}^2_j}{\text{min} } 
& &(T_j-\alpha_1\tilde{t}-\alpha_2\tilde{T}_j^2)^2\\
\iff & \underset{\tilde{t} \in \mathbb{Z}_{p_1}}{\text{min} }
& &\bigg(T_j-\alpha_1\tilde{t}-\alpha_2\bigg[\frac{T_j-\alpha_1\tilde{t}}{\alpha_2}\bigg]_{\mathbb{Z}_{p_2}}\bigg)^2
\end{aligned}
\end{equation}
Because of its high computational cost, this method should be reserved to small tensors only (convolutional kernels).
To emphasize the rational of the proposed line search optimization procedure over the alternating approach, we show the highly non-convex function of $\alpha_1$ and $\alpha_2$ for the INT4 setting in Figure \ref{fig:2D_LS}.
In our experiments, dual line-search approach improves the MSE by 5x in average over the tested dual layers.

Layers with high MSE that are approximated using multiple quantized tensors obviously require more parameters and computations, so a trade-off between accuracy and performance can be established.
Defining speedup is not trivial since it is highly dependent on the hardware design. 
Modern GPUs can double the TOPS performance at INT4 precision, since multiple \emph{small} low precision matrix multiplications can be performed in a distributed fashion.
In order to measure the increase in storage and computations we define the compression ratio for analysis of the framework.
The compression ratio $0<\text{CR}\leq1$ is defined as the ratio of the quantized model (weights and scaling factors) size to the FP32 model size.
In this work, difficult layers in the INT4 setting are approximated with the dual method only ($n=2$).
Performance of the proposed algorithm as well as the corresponding compression ratios are summarized in Table \ref{table:dual}.
\begin{table}
\begin{center}
\begin{tabular}{|p{1.75cm}|p{1.1cm}|p{1.1cm}|p{1.1cm}|p{1.1cm}|}
\hline
Architecture & Original & \ OMSE &\ \ Dual& \ \ \ CR\\
\hline\hline
Alexnet & 56.624\%      & 46.892\%  &\textbf{54.408}\%  		 &0.1256 \\
VGG16bn & 73.476\%      & 65.414\%   &\textbf{66.932}\%         &0.125 \\
Inception v3 & 76.226\% & 22.028\%  &\textbf{51.642}\%       	 &0.1263 \\
Resnet18 & 69.644\%     & 56.688\%  &\textbf{64.034}\%  		 &0.1318 \\
Resnet50 & 76.012\%     & 67.356\%  &\textbf{70.060}\%  		 &0.1261 \\
Resnet101 & 77.314\%     & 65.066\%  &\textbf{71.492}\% 		 &0.1261 \\
SqueezeNet & 58.0\%     & 32.262\%  &\textbf{53.466}\%       	 &0.1493 \\
DenseNet & 74.472\%     & 59.012\%  &\textbf{64.400}\%       	 &0.1432 \\
\hline
\end{tabular}
\end{center}
\caption{Top1 accuracy using INT4 optimal MSE quantization of weights (FP32 activations) and INT4 dual MSE (Dual) for $\tau=8\cdot10^{-5}$. The last column defines the compression ratio (CR) induced by the dual method. CR of regular (non-dual) INT4 linear quantization is $\sim 0.125$.}
\label{table:dual}
\end{table}

\begin{table}
\begin{center}
\begin{tabular}{|l|c|c|c|c|}
\hline
Architecture & Dual & OMSE+Opt. & Dual+Opt.\\
\hline\hline
Alexnet &      54.408\%  &53.306\% & \textbf{55.226} \% \\
VGG16bn &      66.932\%  &72.294\%      & \textbf{72.576}\% \\
Inception v3 & 51.642\%  &73.656\%      & \textbf{74.790}\% \\
Resnet18 &     64.034\%  &67.120\% & \textbf{68.806}\% \\
Resnet50 &     70.060\%  &74.672\% & \textbf{74.976}\% \\
Resnet101 &     71.492\%  &76.226\% & \textbf{76.402}\% \\
SqueezeNet &   53.466\%    &54.514\%      & \textbf{56.248}\% \\
DenseNet &     64.400\%  &71.730\%      & \textbf{73.600}\% \\
\hline
\end{tabular}
\end{center}
\caption{Top1 accuracy using FP32 activations and INT4 dual MSE quantization of weights (Dual) and dual MSE + refined scaling factors (Dual+Opt.) with $\tau=8\cdot10^{-5}$. 
As ablation study we present the refinement of the OMSE method without dual quantization (OMSE+Opt).
The calibration set contains five hundred images from the ImageNet validation set and 25 epochs.}
\label{table:gamma}
\end{table}

\begin{table*}[t]
\begin{tabular}{|l|c|c|c|c|c||c|c|c|c|c|}
\hline
Architecture                    	& Original      & Baseline              & KLD            		  &OMSE   	& Dual (CR)								 & $\text{ACIQ}^{*}$                         &$\text{OMSE}^{*}$          &$\text{Dual}^{*}$                                 \\
\hline\hline
Alexnet                        	& 56.624\%      & 38.616\%				& 44.23\%			& \textbf{48.908}\%			&  \textbf{54.552}\% (0.195) 			&-             & \textbf{49.122}\%                      	&\textbf{54.994}\%                          \\
Alexnet (offset)               	& 56.624\%      & 51.774\%				& 52.948\%			& \textbf{53.286}\%   &  \textbf{55.522}\% (0.195)            	    &52.304\%          & \textbf{53.998}\%      				&\textbf{55.508}\%  	                          \\
\hline
VGG16bn                        	& 73.476\%      & 33.276\%				&53.686\%			& \textbf{62.168}\%            &  \textbf{68.120}\% (0.127)       		&-           & \textbf{67.726}\%                     	& \textbf{68.334}\%                                  \\
VGG16bn (offset)              	& 73.476\%      & 58.832\%				& 64.336\%			 & \textbf{67.198}\%    &  \textbf{71.478}\%  (0.127)      		&67.716\%            & \textbf{71.260}\%                    	&\textbf{71.260}\%  		                                  \\
\hline
Inception v3                  	& 76.226\%      & 4.042 \%				& 23.658 \%			& \textbf{40.916}\%            &  \textbf{66.176}\%  (0.154)         	 &-             & \textbf{43.184}\%       				&   \textbf{68.608}\%                        \\
Inception v3 (offset) 			& 76.226\%      & 57.516 \%				& 64.504\%			& \textbf{67.964}\%     &  \textbf{73.060}\% (0.151)     	          &59.826\%      & \textbf{69.528}\%                      	&   \textbf{74.486}\%                                  \\
\hline
Resnet18                       	& 69.644\%		& 48.37\%				& 56.728\%			 & \textbf{61.268}\%           &  \textbf{66.522}\% (0.150)           &-              & \textbf{63.744}\%      				 &      \textbf{66.628}\%                      \\
Resnet18 (offset)    			& 69.644\%      & 63.106\%				& 64.486\%			& \textbf{64.992}\%      &  \textbf{68.380}\%  (0.148)        	   &65.694\%        & \textbf{67.508}\%                       &       \textbf{68.316}\%                       \\
\hline
Resnet50                  		& 76.012\%      & 40.786\%				& 57.57\%			& \textbf{64.878}\%           &  \textbf{70.368}\% (0.129)        	&-               & \textbf{66.562}\%         				 &      \textbf{70.202}\%                         \\
Resnet50 (offset)      			& 76.012\%      & 65.338\%				& 68.328\%			& \textbf{71.274}\%     &  \textbf{73.252}\% (0.126)          	    &71.362\%        & \textbf{73.392}\%                      &	    \textbf{73.392}\%                              \\
\hline
Resnet101                      	& 77.314\%      & 36.494\%				& 58.744\%			& \textbf{65.316}\%            &  \textbf{70.770}\% (0.129)        	 &-             & \textbf{65.350}\%       				&\textbf{70.806}\%                                \\
Resnet101 (offset)             	& 77.314\%      & 65.552\%				& 71.412\%			& \textbf{72.750}\%     &  \textbf{74.266}\% (0.126)                  &69.544\%        & \textbf{74.332}\%       				&         \textbf{74.332}\%                        \\
\hline
SqueezeNet                     	& 58.0\%        & 3.282\%				& 9.582\%			& \textbf{18.630}\%            &  \textbf{52.382}\% (0.216)            &-           & \textbf{20.448}\%       					 &          \textbf{52.430}\%                               \\
SqueezeNet (offset)   			& 58.0\%        & 24.97\%				& 35.806\%			& \textbf{39.820}\%            &  \textbf{56.150}\% (0.203)        	 &-           & \textbf{42.026}\%                        &            \textbf{56.284}\%                              \\
\hline
DenseNet                       	& 74.472\%      &47.808\%				& 64.062\%			& \textbf{65.032}\%            &  \textbf{67.952}\% (0.132)         	 &-       & \textbf{67.558}\%                				&       \textbf{68.048}\%                             \\
DenseNet (offset)     			& 74.472\%      & 67.676\%				& 69.578\%			 & \textbf{70.118}\%            	&  \textbf{72.282}\% (0.132)          &-      & \textbf{72.304}\%                             &       \textbf{72.310}\%                \\
\hline
\end{tabular}
\caption{Top1 accuracy using INT4 MSE quantization of activations, INT8 weights uniformly quantized and dual MSE ($\tau=8\cdot10^{-5}$) compared with several methods.
The baseline is defined as in \cite{jacob2017quantization} with clipping to the mean maximum and minimum range of the calibration set.
KLD denotes the saturation technique based on Kullback-Leibler divergence metric proposed in \cite{migacz}.
In ACIQ framework  \cite{banner2018aciq} the first layer of the models is not quantized and unsigned representation is used. ACIQ results are taken from the paper.
We provide similar to ACIQ experiment with our methods ($\text{OMSE}^{*}$ and $\text{Dual}^{*}$) wherein we do not quantize the input layer (if first layer is the only key layer, $\text{OMSE}^{*}$ and $\text{Dual}^{*}$ yield same results).
The calibration set is composed of 250 images for all the methods.
}
\label{table:activ}
\end{table*}
\subsection{Scaling Factors Refinement}

One issue with the quantization method proposed above is the rigidity of the quantization mapping performed axiomatically according to a given metric.
In order to tackle this problem without the need for full retraining and without requiring gradients of non differentiable functions, we propose a post quantization adjustment of the scaling factors of the quantized NN weights.
Given calibration data (unlabeled), we refine the scaling factors to better approximate the full precision model. 
We seek to optimize the (re)scaling factor $\gamma$ defined as $\hat{T}(\gamma)=\gamma\alpha\tilde{T}$, with $\tilde{T}$ being the tensor  quantized axiomatically.
Consequently, the saturation threshold $\alpha$ is approximated for optimal reconstruction separately, using $L_2$ metric. The rescaling factor $\gamma$ is optimized afterwards in a data driven way such that
\begin{equation}
\begin{aligned}
\hat{T}(\gamma)=\gamma\bigg(\alpha\bigg[\frac{T}{\alpha} \bigg]_{\mathbb{Z}_{p}}\bigg).
\end{aligned}
\end{equation}
Assuming $f(X,\{W_l\}_l)$ is the NN mapping function, we seek to minimize
\begin{equation}
\begin{aligned}
& \underset{\gamma_l=\{\gamma_{lk}\}_{lk}}{\text{min}}
& & \sum_i^M\|f(X_i,\{W_l\}_l)-f(X_i,\{\hat{W_l}(\gamma_l)\}_l)\|^2_F,
\end{aligned}
\end{equation}
where $M$ is the size of the calibration set.
The advantage of this approach is that the number of optimized values is small and equal to the number of convolutional kernels. 
At contrary to training methods, this approach is fully \emph{differentiable} avoiding sub-gradient definitions. Thus, the optimization can be conducted very efficiently using popular stochastic gradient descent methods on a \emph{small} calibration set.
Also, only a few optimization steps are required for both fast deployment and better generalization.
Results of the refinement procedure, and its influence on accuracy improvement are presented in Table \ref{table:gamma}. 
These  experiments show that the refinement procedure can improve the accuracy by up to to 23 percent.
We also provide ablation study to demonstrate the usefulness of the refinement stage in low compression ratio tasks where no dual layers are allowed.


\subsection{Activations Quantization}\label{sec:activations}

In contrast to weights quantization, the activations quantization should be performed for each image on the fly. Obviously, optimal scaling factors of activations cannot be computed for every image. 
Thus, the quantization parameters are generally approximated using relatively small calibration set \cite{migacz,banner2018aciq} where saturating the activation values to a given threshold helps in improving accuracy.
The saturation problem is usually solved by approximating the activations \emph{distribution} \cite{migacz,banner2018aciq}. These methods are less sensitive to outliers with potential important impact on the underlying dot product.
Here, we follow the same MSE minimization approach, wherein the parameters are obtained as described in Section \ref{section:optimal_mse}.
Thereby, given a small calibration set (few hundred data samples) we collect activations at different layers of the network, and seek for optimal saturation factor.
Given $X_l\in \mathbb{R}^{d,c_l,w_l,h_l}$ the $l^{th}$ layer activations with $d$ the size of the calibration set, we solve
  \begin{equation}
  \label{eq:act1}
\begin{aligned}
& \underset{\beta_1}{\text{min}}
& & \|X_l -\beta_1\bigg[\frac{X_l}{\beta_1}\bigg]_{\mathbb{Z}_{p}}\|^2_F.
\end{aligned} 
\end{equation}

\begin{table*}[ht]
\begin{center}
\begin{tabular}{| *{9}{c|}}
\hline
Architecture &  \multicolumn{2}{c|}{Original }  & \multicolumn{2}{c|}{Our} & \multicolumn{2}{c|}{Our+offset} & CR(W,A)			& \% Dual\\
\hline\hline
Alexnet 		& 56.624\%	&79.056\%		& 53.632\%	&77.244\%			&  \textbf{54.476}\% & \textbf{77.846}\% &(0.125,0.195)		&25\%\\
VGG16bn 		& 73.476\%	&91.536\%		& 67.492\%	&88.016\%			&  \textbf{70.658}\% & \textbf{90.136}\% &(0.125,0.127)  &6.25\%\\
Resnet18		 & 69.644\%	&88.982\%		&65.456\%	&86.630\%			&  \textbf{67.416}\% & \textbf{87.716}\% &(0.126,0.148)  &23.80\%\\
Resnet50		 & 76.012\%	&92.934\%		& 69.370\%	&89.204\%			&  \textbf{72.602}\% & \textbf{90.852}\% &(0.126,0.129)  &3.70\%\\
Resnet101 	& 77.314\%	&93.556\%		& 69.700\%	&89.686\%			&  \textbf{73.602}\% & \textbf{91.526}\% &(0.126,0.128) &1.90\%\\
Inception v3	 & 76.226\%	&92.984\%		& 64.572\%	&85.852\%			&  \textbf{71.606}\% & \textbf{90.470}\% &(0.126,0.154) &11.57\%\\
SqueezeNet	 & 58.184\%	&80.514\%		& 50.722\%	&74.634\%			&  \textbf{55.358}\% & \textbf{78.482}\%  &(0.149,0.216) &68\%\\
DenseNet 	& 74.472\%	&91.974\%		& 66.832\%	&87.518\%			&  \textbf{71.558}\% & \textbf{90.532}\% &(0.143,0.132) &2.47\%\\
\hline
\end{tabular}
\end{center}
\caption{Performance of the proposed framework for INT4 weights and activations obtained using $\tau=8\cdot 10^{-5}$. Compression ratios of both the weights and the activations and the percentage of dual layers are provided in the last two columns, respectively. The mean top-1 decay is of 6.7\% and 3\% for the regular and offset (unsigned) versions, respectively.
Similarly, mean top-5 decay is of 4\% and 1.7\% respectively.}
\label{table:main_res}
\end{table*}
For key layers, optimal approximation using multiple quantizations as described in Section \ref{section:multiple} suffers from strong overfitting. 
In order to approximate better, and to generalize at the same time, we obtain the optimal parameters by quantizing the \emph{residual} from the first approximation, such that for a given activation $X$ we have
  \begin{equation}
  \label{eq:act11}
\begin{aligned}
X\approx\hat{X}&=\beta_1\tilde{X}_1+\beta_2\tilde{X}_2\\
&=\beta_1\bigg[\frac{X}{\beta_1}\bigg]_{\mathbb{Z}_p}+\beta_2{\bigg[\frac{X-\beta_1\tilde{X}_1}{\beta_2}\bigg]_{\mathbb{Z}_p}}.
\end{aligned}
\end{equation}
We obtain $\beta_1$ from eq.(\ref{eq:act1}), followed by obtaining  $\beta_2$ from eq.(\ref{eq:act11}).
This procedure is similar to the alternating approach described in Algorithm \ref{alg:dual} with one iteration only.
\begin{figure}
\begin{center}
   \includegraphics[width=1\linewidth]{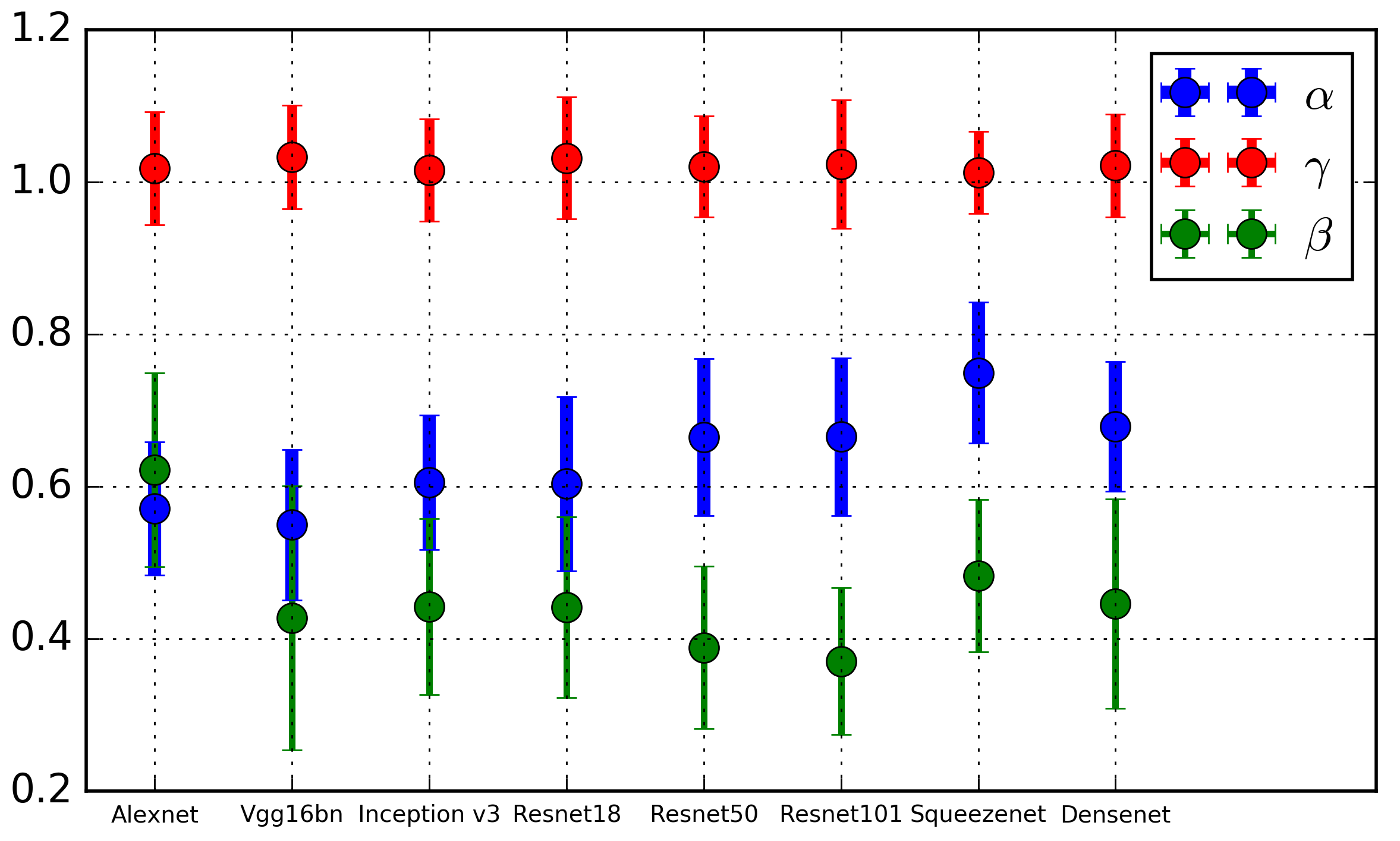}
   \caption{Mean and standard deviation of the set of scaling factors of the weights ($\alpha$), the activations ($\beta$) on the calibration set (not dual) and the refined factors ($\gamma$), all normalized by the maximum value of the tensor they approximate.}
   \label{fig:satur}
   \end{center}
\end{figure}
Similar to the network weight compression ratio,  we define the activations compression ratio as the ratio between the size of all the compressed activations and the size of all the original activations.

Table \ref{table:activ} presents the comparison of the variants of the proposed activation quantization method with several existing methods. The line-search method iterates over 50 samples only.
Statistics of thresholding factors for different models are presented in Figure \ref{fig:satur} where we can observe the severe saturation of the MMSE method.

\section{Experiments}
We evaluate the proposed framework on several popular architectures in order to demonstrate its robustness to aggressive quantization. We also evaluate quantization of non over-parameterized models such as SqueezeNet or DenseNet that are much more sensitive to quantization and are usually not analyzed in the NN quantization literature.
We consider different INT4 quantization scenarios such as unsigned (with offset) and signed representations. We present also the accuracy-compression ratio trade off analysis.
The framework has been implemented using the Pytorch library with its pretrained models. 

The MSE grid search iterates over 500 sampling points for the weights and 50 for the activations. The scaling factors refinement stage requires 25 epochs over 500 randomly sampled images from the validation set. 
The calibration set used for the quantization of activations is comprised of 250 images.
\emph{All} the layers are quantized, including the network input.
Also, it is important to notice that there is currently no existing framework for linear INT4 only quantization of weights and activations for comparison in the inference setting.
The summary of the accuracy loss and the compression ratios is presented in Table \ref{table:main_res}.
In Table \ref{table:tradeoff} we show how modification of the MSE threshold $\tau$ influences the accuracy-compression ratio trade-off.
\begin{table}
\begin{center}
\begin{tabular}{|p{1.6cm}|p{1.1cm}|p{1.1cm}|p{1.75cm}|p{1.3cm}|}
\hline
Architecture &  Original  & Our & CR(W,A) & Our +offset\\
\hline\hline
\multirow{ 3}{*}{Alexnet}	&\multirow{ 3}{*}{56.624\%}		& 49.570\%	&(0.125,0.178) &52.274\%\\
& &54.186\%	&(0.127,0.206) 	&54.936\%\\
& &54.720\%	&(0.130,0.245)	&56.068\%\\
\hline
\multirow{ 3}{*}{Resnet50}	&\multirow{ 3}{*}{76.012\%}		& 69.472\%	&(0.126,0.129) &72.530\%\\
& &73.078\%	&(0.137,0.156) 	&74.826\%\\
& &74.336\%	&(0.154,0.203)	&75.198\%\\
\hline
\multirow{ 3}{*}{Inception v3}	&\multirow{ 3}{*}{76.226\%}		& 61.856\%	&(0.126,0.146)&71.662\%\\
& &74.036\%	&(0.150,0.215) &75.354\%	\\
& &74.766\%	&(0.186,0.238) &75.870\%	\\
\hline
\multirow{ 3}{*}{DenseNet}	&\multirow{ 3}{*}{74.472\%}		& 65.526\%	&(0.126,0.132)&70.730\%\\
& &70.716\%	&(0.155,0.154) 	&73.114\%\\
& &73.382\%	&(0.226,0.241)	&74.116\%\\
\hline
\end{tabular}
\end{center}
\caption{Trade-off analysis of the proposed method without and with offset for three thresholding  values $\tau =\{10,2,0.9\}\cdot 10^{-5}$. The compression ratio refers to the method without offset.}
\label{table:tradeoff}
\end{table}

\section{Conclusion}
In this paper we introduced an efficient and accurate MSE-based low-bit precision quantization framework for neural networks.
Our approach deploys hardware-aware partitioning of the network parameters, and the refinement of high MSE layers, using quantized filter banks.
Given a small calibration set, we further refine the quantization scaling factors for better approximation of the original model.
We also provide a framework for the quantization of the network activations, wherein we propose a method of residual quantization for improved approximation of the most sensitive layers.
The proposed approach can be adjusted to any desired precision for constrained hardware deployment, according to the inherent compression-complexity trade-off of the method.
The framework allows fast and efficient deployment of pretrained models, producing a new state-of-the-art INT4 inference quantization results.

\clearpage
\newpage
{\small
\bibliographystyle{ieee}
\bibliography{egbib}
}

\end{document}